\documentclass{article}

\usepackage{arxiv}

\usepackage{amsmath,multirow}
\usepackage{tabularx,booktabs,makecell,subcaption,setspace}
\usepackage{siunitx}
\usepackage[normalem]{ulem}
\useunder{\uline}{\ul}{}
\usepackage[utf8]{inputenc} % allow utf-8 input
\usepackage[T1]{fontenc}    % use 8-bit T1 fonts
\usepackage{hyperref}       % hyperlinks
\usepackage{url}            % simple URL typesetting
\usepackage{booktabs}       % professional-quality tables
\usepackage{amsfonts}       % blackboard math symbols
\usepackage{nicefrac}       % compact symbols for 1/2, etc.
\usepackage{microtype}      % microtypography
\usepackage{lipsum}		% Can be removed after putting your text content
\usepackage{graphicx}
\usepackage[numbers]{natbib}
\usepackage{doi}

\title{Beat-ssl: Capturing Local ECG Morphology through Heartbeat-level Contrastive Learning with Soft Targets}

\author{
    Muhammad Ilham Rizqyawan$^{\star \diamond}$ \qquad Peter Macfarlane$^{\dagger}$ \qquad Stathis Hadjidemetriou$^{\ddagger}$  \qquad Fani Deligianni$^{\star}$\thanks{Corresponding author: fani.deligianni@glasgow.ac.uk --- Accepted at ISBI 2026}\\
	$^{\star}$School of Computing Science, University of Glasgow, UK\\
        $^{\diamond}$Research Center for Smart Mechatronics, National Research and Innovation Agency, Indonesia\\
        $^{\dagger}$School of Health \& Wellbeing, University of Glasgow, UK\\
        $^{\ddagger}$Department of Information Technologies, University of Limassol, Cyprus\\
}

\date{}

\renewcommand{\shorttitle}{Beat-SSL}

\hypersetup{
pdftitle={Beat-ssl: Capturing Local ECG Morphology through Heartbeat-level Contrastive Learning with Soft Targets},
pdfsubject={q-bio.NC, q-bio.QM},
pdfauthor={Muhammad I.~Rizqyawan, Peter~MacFarlane, Stathis~Hadjidemetriou, Fani~Deligianni},
pdfkeywords={ECG, semi-supervised learning, contrastive learning, ECG segmentation, multilabel classification},
}

\begin{document}
\maketitle
\begin{abstract}
	Obtaining labelled ECG data for developing supervised models is challenging. Contrastive learning (CL) has emerged as a promising pretraining approach that enables effective transfer learning with limited labelled data. However, existing CL frameworks either focus solely on global context or fail to exploit ECG-specific characteristics. Furthermore, these methods rely on hard contrastive targets, which may not adequately capture the continuous nature of feature similarity in ECG signals. In this paper, we propose Beat-SSL, a contrastive learning framework that performs dual-context learning through both rhythm-level and heartbeat-level contrasting with soft targets. We evaluated our pretrained model on two downstream tasks: 1) multilabel classification for global rhythm assessment, and 2) ECG segmentation to assess its capacity to learn representations across both contexts. We conducted an ablation study and compared the best configuration with three other methods, including one ECG foundation model. Despite the foundation model's broader pretraining, Beat-SSL reached 93\% of its performance in multilabel classification task and surpassed all other methods in the segmentation task by 4\%.
\end{abstract}

% keywords can be removed

\section{Introduction}
\label{sec:intro}
Recent studies have demonstrated the effectiveness of machine learning models to satisfactorily undertake 12-lead ECG analysis, even for low-prevalence diagnoses such as congenital heart disease \cite{alkanRiemannianPredictionAnatomical2024, chenCongenitalHeartDisease2024}. However, training deep learning models requires large amounts of labelled ECG data, which remains challenging to obtain even at a national scale. Foundation models pretrained on unlabelled data offer a promising solution, enabling generalisation across diverse tasks through transfer learning with minimal labelled samples \cite{hanFoundationModelsElectrocardiogram2024,mckeenECGFMOpenElectrocardiogram2024}. 

Contrastive learning (CL) has emerged as an effective pretraining approach that leverages unlabelled data by learning to distinguish similar and dissimilar samples \cite{chenSimpleFrameworkContrastive2020,heMomentumContrastUnsupervised2020}. While several CL frameworks have been proposed for biomedical signal analysis, many are influenced by computer vision paradigms or are designed for general time series \cite{yueTS2VecUniversalRepresentation2022}, failing to exploit ECG specific characteristics.
% \cite{,gaoSimCSESimpleContrastive2022,izacardUnsupervisedDenseInformation2022}

The 12-lead ECG encodes 3D cardiac activity across temporal cycles through its spatio-temporal structure, providing rich information for learning robust representations. 3KG \cite{gopal3KGContrastiveLearning2021} leverages this by transforming ECG signals into the vectorcardiography (VCG) domain, a 3D representation of cardiac activity. This is used as an augmentation strategy but considers only global context. Oh et al \cite{ohLeadagnosticSelfsupervisedLearning2022} proposed capturing both global context using Contrastive Multi-segment Coding (CMSC) \cite{kiyassehCLOCSContrastiveLearning2021} (also used by 3KG) and local context using Wave2vec \cite{baevskiWav2vec20Framework2020}. The ECG-FM foundational model adopts this dual context pretraining approach \cite{mckeenECGFMOpenElectrocardiogram2024}. However, these methods overlook that ECG signals comprise multiple heartbeats with varying intervals and durations, which represent fundamental units of cardiac activity that could provide richer contrastive cues.

% ============= Full paragraph VERSION =============
In this paper, we propose an ECG CL framework that considers both global and local contexts by leveraging heartbeat-level contrasting for local context representation. First, we use the 12-lead ECG as input, which is transformed into the 3D VCG domain and augmented the signal using the 3KG strategy \cite{gopal3KGContrastiveLearning2021}. Subsequently, we apply contrastive learning in both global and local contexts. We introduce two soft contrastive strategies that utilised ECG feature similarity as a continuous target between 0 and 1, moving beyond binary positive-negative pairs. Finally, we validate the model on two downstream tasks: multilabel classification for global context evaluation and ECG wave segmentation for local context evaluation.

\section{Methods}
\label{sec:methods}
\begin{figure*}[t]
    \centering
    \includegraphics[width=1\linewidth]{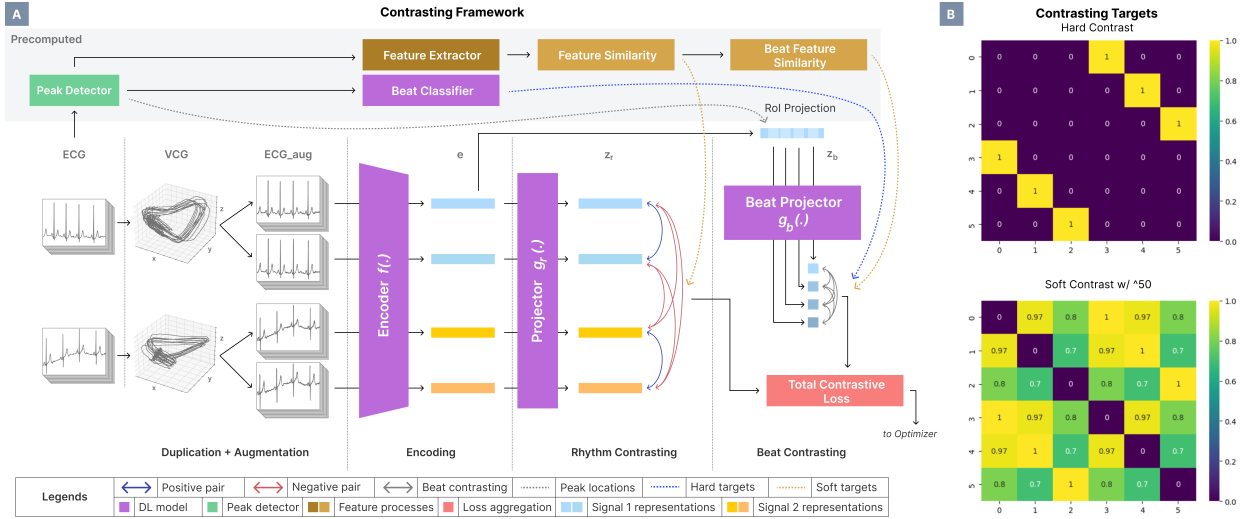}
    \caption{A) General structure of the proposed framework with both rhythm and beat level contrasting. The hard targets for beat level are obtained from the beat-classifier (blue line). The soft targets are obtained from features similarity (brown lines). B) Comparison between hard targets (top), which use binary values either 0 or 1, and soft targets (bottom) which use continuous values ranging from 0 to 1.}
    \label{fig:generalframework}
\end{figure*}

Within this framework, we aim to leverage the unique characteristics of ECG signals. ECGs can be analysed at two levels: the rhythm level, which encompasses several heartbeats, and the morphological level, which captures beat-specific characteristics. Following  Oh et al. \cite{ohLeadagnosticSelfsupervisedLearning2022}, we refer to these as the global and local contexts, respectively. Contrasting at the beat level is essential for enabling the model to differentiate individual beat morphologies by identifying the characteristics of specific waves (P-wave, QRS-complex, and T-wave). Therefore, we propose a framework that performs contrastive learning at both the rhythm and beat levels. The overall concept of the framework is illustrated in Fig. \ref{fig:generalframework}A.

\subsection{Global Context: Rhythm Contrasting}
\label{sec:method_inst_contrast}

Our framework extracts useful representations from unlabelled data by teaching the model to recognise that different augmented versions of the same signal are similar, while different signals are dissimilar \cite{chenSimpleFrameworkContrastive2020}. 
Formally, given an ECG signal $\mathbf{X} \in \mathbf{R}^{L \times D}$, where $L$ is the number of leads (12 in this case) and $D$ is the temporal dimension (5000 samples at 500 Hz for 10 seconds), we generate two augmented views, $\tilde{\mathbf{X}}_i$ and $\tilde{\mathbf{X}}_j$, through stochastic data augmentation operations.

% We adopt the 3KG \cite{gopal3KGContrastiveLearning2021} augmentation strategy, which operates in the vectorcardiography (VCG) domain. The transformation pipeline can be expressed as:
We adopt the 3KG \cite{gopal3KGContrastiveLearning2021} augmentation strategy, which operates in the VCG domain. The transformation pipeline can be expressed as:

\begin{equation}
\tilde{\mathbf{X}} = \mathbf{M}^{-1}_{\text{Kors}} \cdot \mathcal{A}(\mathbf{M}_{\text{Kors}} \cdot \mathbf{X}) + \boldsymbol{\epsilon}
\end{equation}

\noindent where $\mathbf{M}_{\text{Kors}} \in \mathbf{R}^{3 \times 12}$ is the Kors transformation \cite{korsReconstructionFrankVectorcardiogram1990} matrix for ECG-to-VCG conversion, $\mathcal{A}(\cdot)$ represents augmentation operations (rotation and scaling) applied in the VCG space, $\mathbf{M}^{-1}_{\text{Kors}}$ is the pseudo-inverse used for VCG-to-ECG reconstruction, and $\boldsymbol{\epsilon} \sim \mathcal{N}(0, \sigma^2)$ is additive Gaussian noise in the ECG domain with noise scale $\sigma = 0.05$. The rotation operation applies a random angle $\theta \in [\ang{-15}, \ang{15}]$  around a randomly selected axis, and scaling applies a factor $s \in [1.0, 1.2]$. The Kors transformation matrix has been shown to yield the highest reconstruction accuracy \cite{vondrakStatisticalEvaluationTransformation2022}. These parameter values were chosen to ensure that the augmentations remains physiologically meaningful and do not distort the signal into pathological forms.

\subsection{Local Context: Heartbeat Contrasting}
\label{sec:method_beat_contrast}
For beat-level analysis, we process individual heartbeats cropped to $n$ samples centred at the R-peak, ensuring that each segment contains a complete heartbeat. However, directly passing these cropped beat signals to the base encoder is not feasible due to mismatched input dimensions. Using a separate encoder is also undesirable, as our goal is to maintain a single encoder throughout pretraining.

To address this, we input the full 12-lead, 10-second ECG signal into the encoder once, as in rhythm-level contrasting. For beat-level contrast, we then extract beat-specific representations by splitting the encoder output based on the projected R-peak positions. This approach is analogous to the Region of Interest (ROI) pooling in Fast R-CNN \cite{girshickFastRCNN2015}. The extracted representations are subsequently passed through a dedicated beat projection head, $g_b(\cdot)$.

At the beat level, we use pseudo-labels as hard contrastive targets for the contrastive loss function. These labels are generated by a separate beat-classifier trained on Lead-II during preprocessing. This classifier was trained on the MIT-BIH Arrhythmia dataset \cite{moodyImpactMITBIHArrhythmia2001}, using the five AAMII superclasses \cite{iso/ansi/aamiANSIAAMIISO2008}.

\subsection{Soft Contrasting}
\label{sec:method_soft_contrast}
We explore the use of continuous similarity measures as contrastive targets. Instead of using binary values (0 or 1) as similarity targets (Fig. \ref{fig:generalframework}.B, top), we employ soft targets ranging between 0 and 1 (Fig. \ref{fig:generalframework}.B, bottom). This approach is conceptually similar to \cite{leeSoftContrastiveLearning2024}, but differs in how the soft targets are computed.

In this work, a set of features is extracted from the ECG signal during preprocessing. Based on these features, we created two versions of the soft targets. For the \textbf{first version} (soft\_1), a soft target matrix of size $2N \times 2N$ is produced by concatenating the feature set $N$ with itself to form a $2N$-sized tensor, followed by computing pairwise cosine similarities. To emphasize high similarity values, the resulting similarities are exponentiated using a power exponentiation factor.
% extracted from Lead II of the ECG signal
% To account for high similarity between features, we scaled the similarity by raising it to a power ranging from 1 to 75.

For the \textbf{second version} (soft\_2), the $p$-norm distance is computed based on the feature set, and the top-$k$ neighbours are identified for each sample. We assign weights based on proximity: for $k=3$, the three nearest neighbours received weights of $1$, $\frac{2}{3}$, and $\frac{1}{3}$, respectively, while all other samples are assigned a weight of zero. These weights are then concatenated to form a $2N \times 2N$ matrix. This approach is similar to \cite{guTransformingLabelefficientDecoding2025}. However, instead of adding $k$ hard positive pairs, we assign soft weights based on the neighbour ranking.

% This approach is similar to \cite{guTransformingLabelefficientDecoding2025}, which identifies the nearest neighbours from the extracted features. However, instead of adding $k$ hard positive pairs, we assign soft weights based on the neighbour ranking.

\subsection{Loss Function}
\label{sec:method_loss}
The generalised form of Normalised Temperature-scaled Cross Entropy (NT-Xent) \cite{chenSimpleFrameworkContrastive2020, sohnImprovedDeepMetric2016, wuUnsupervisedFeatureLearning2018, oordRepresentationLearningContrastive2019} is used as the contrasting loss function. For a mini-batch of $N$ samples, $2N$ samples are generated by creating two views of each sample. The loss for a positive pair $(i, j)$ is defined as:

% \[l_{i,j} = -log\frac{exp(sim(z_{i}, z_{j})/\tau)}{\sum_{k\neq i}^{2N}exp(sim(z_{i}, z_{k})/\tau)}\]

% \noindent where $z$ is the 128-dimensional vector produced by the projection heads, $sim(i, j)$ denotes the cosine similarity between $i$ and $j$, and $\tau$ is the temperature parameter. Unless otherwise stated, outputs from both the instance and beat-levels projection head $g(.)$ and $g\_b(.)$ are passed to the same loss function. The total loss is computed as follows:

% \[l_{total} = \lambda.l_{i}+(1-\lambda).l_{b}\]

% \noindent where $l_{i}$ and $l_{b}$ denotes the instance loss and beat loss respectively, and $\lambda$ represents weighting factor between them.

% To compute the soft contrastive loss, we used a generalised form of NT-Xent loss:

\begin{equation}
l_{i,j} = -\sum_{k\neq i}^{2N}w_{i,k}.log\frac{exp(sim(z_{i}, z_{j})/\tau)}{\sum_{m\neq i}^{2N}exp(sim(z_{i}, z_{m})/\tau)}
\end{equation}

\noindent where $z$ is the 128-dimensional vector produced by the projection heads, $sim(i, j)$ denotes the cosine similarity between $i$ and $j$, $\tau$ is the temperature parameter, and $w_{i,k}$ is the contrastive target.

% The total loss is computed as $l_{total} = \lambda.l_{r}+(1-\lambda).l_{b}$, where $l_{r}$ and $l_{b}$ denote the rhythm loss and beat loss, respectively, and $\lambda$ is the weighting factor between them.

% \noindent where $w_{i,k}$ is the soft target value between 0 and 1, as described above.

\subsection{Experiments}
\label{sec:method_experiment}
For the experiment, we conducted an ablation study by pretrained the model with 14 combinations, and performed two downstream tasks. We compared the best configuration with other frameworks, including TS2Vec\cite{yueTS2VecUniversalRepresentation2022}, and Domain-SSL\cite{guTransformingLabelefficientDecoding2025}. We also included ECG-FM\cite{mckeenECGFMOpenElectrocardiogram2024}, an ECG foundation model pretrained using the framework of Oh et al.\cite{ohLeadagnosticSelfsupervisedLearning2022}.

\subsubsection{Pretraining}
\label{sec:exp_pre}
For pretraining, the PTB-XL dataset \cite{wagnerPTBXLLargePublicly2020} is used without labels. The dataset contains 21,837 records from 18,885 patients. The dataset provides stratified fold assignments to reduce bias and prevent data from the same patient appearing in multiple folds. The input consists of 10-second, 12-lead ECG signals sampled at 500 Hz. For pretraining, we used folds 1 to 8.
%For pretraining, we used folds 1 to 8. Unless otherwise stated, we used AdamW optimizer with a learning rate of $1\times10^{-4}$, $\tau$ = 0.25, $\lambda$ = 0.5, batch size = 256, noise scale = 0.05, max scale = 1.2, and a VCG rotation range = -15\textdegree to 15\textdegree.

\subsubsection{Downstream tasks}
\label{sec:exp_ds}
\textbf{Multi-label classification.} For the first downstream task, the superdiagnostic task from the PTB-XL dataset was used. It includes five classes: Normal (NORM), Conduction Disturbance (CD), Myocardial Infarction (MI), Hypertrophy (HYP), and ST/T Change (STTC). Since multiple classes can be assigned to a single sample, this constitutes a multi-label classification problem. We used folds 1 to 8 for training, fold 9 for validation, and fold 10 for testing.

Linear probing was performed by freezing the pre-trained encoder and projecting its output through a single linear layer. Prior to projection, we applied 1D max-pooling (kernel size = 4) operation along the temporal axis to improve computational efficiency. Binary Cross-Entropy was used as the loss function, with Adam as the optimizer.
%For evaluation, we report F1-score, AUROC, Precision, and Recall.

\textbf{ECG Segmentation.} For the second downstream task, the LUDB \cite{kalyakulinaLUDBNewOpenAccess2020} delineation dataset was used. It contains 200 records from 200 subjects, with fiducial points (start, peak, and end) for P-wave, QRS-complex, and T-wave across all 12 leads as the label. We computed the average fiducial points across the 12 leads and converted them into masks for each wave. This is formulated as a segmentation task with sample-wise classification. In this setup, we freeze the pre-trained encoder and construct a decoder based on a mirrored architecture of the encoder, with a single output channel. Cross-entropy is used as the loss function, and Adam is employed as the optimizer. During evaluation, to avoid false positives from missing boundary labels in LUDB, we restricted metrics to samples [500–4500].

%resulting in label data of shape [1, 5000] per record, where 5000 corresponds to the same number of samples in the ECG signal. We then treated this as a segmentation task classifying each sample in the signal.

%F1-score, Precision, and Recall were reported as evaluation metrics.

% It is worth noting that in the LUDB dataset, fiducial points are missing for all waves in the first and last heartbeats, the P-wave in the second, and the T-wave in the second-to-last heartbeat. Most waves appear between samples 750 and 4250. The regions from samples 250 to 750 and 4250 to 4750 contain missing labels; however, due to high heart rates in some records, second and second-to-last beats also appear in these regions. Including these areas in the evaluation could lead to high false positives due to the missing labels. Therefore, during evaluation, we restricted metric calculation to samples [500, 4500].

% It is worth noting that the LUDB dataset contains missing labels at the start and the end of all records. To reduce the number of false positives during evaluation, we restricted metric calculation to samples in the range [500, 4500].

% ============= WILCOXON VERSION =============
\begin{figure*}[!ht]
  \centering
  \begin{subcaptionbox}{PTB-XL F1-score\label{fig:signi_ptb_f1}}[0.32\textwidth]
    {\includegraphics[width=\linewidth]{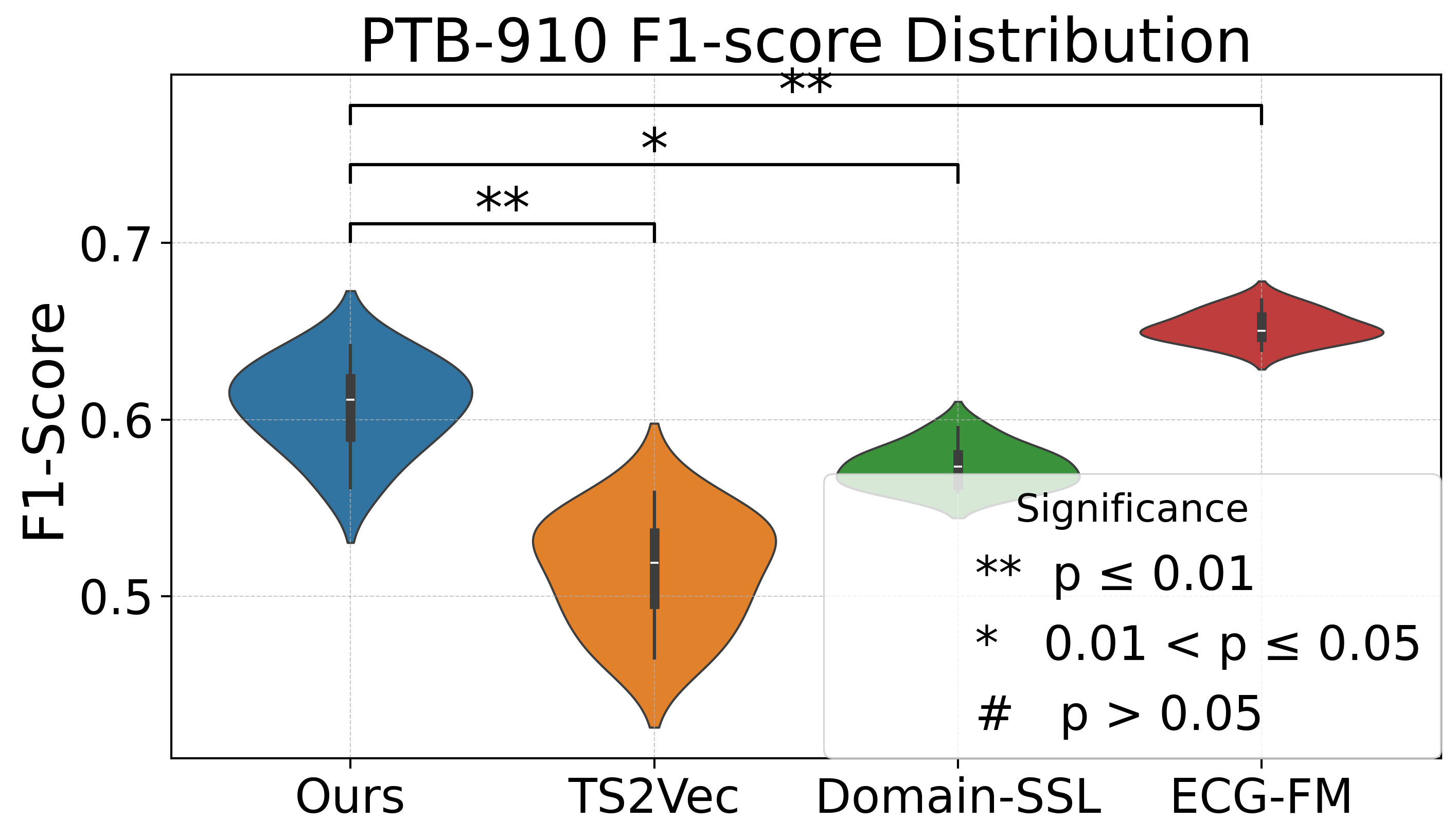}}
  \end{subcaptionbox}
  \hfill
  \begin{subcaptionbox}{LUDB F1-score\label{fig:signi_lu_f1}}[0.32\textwidth]
    {\includegraphics[width=\linewidth]{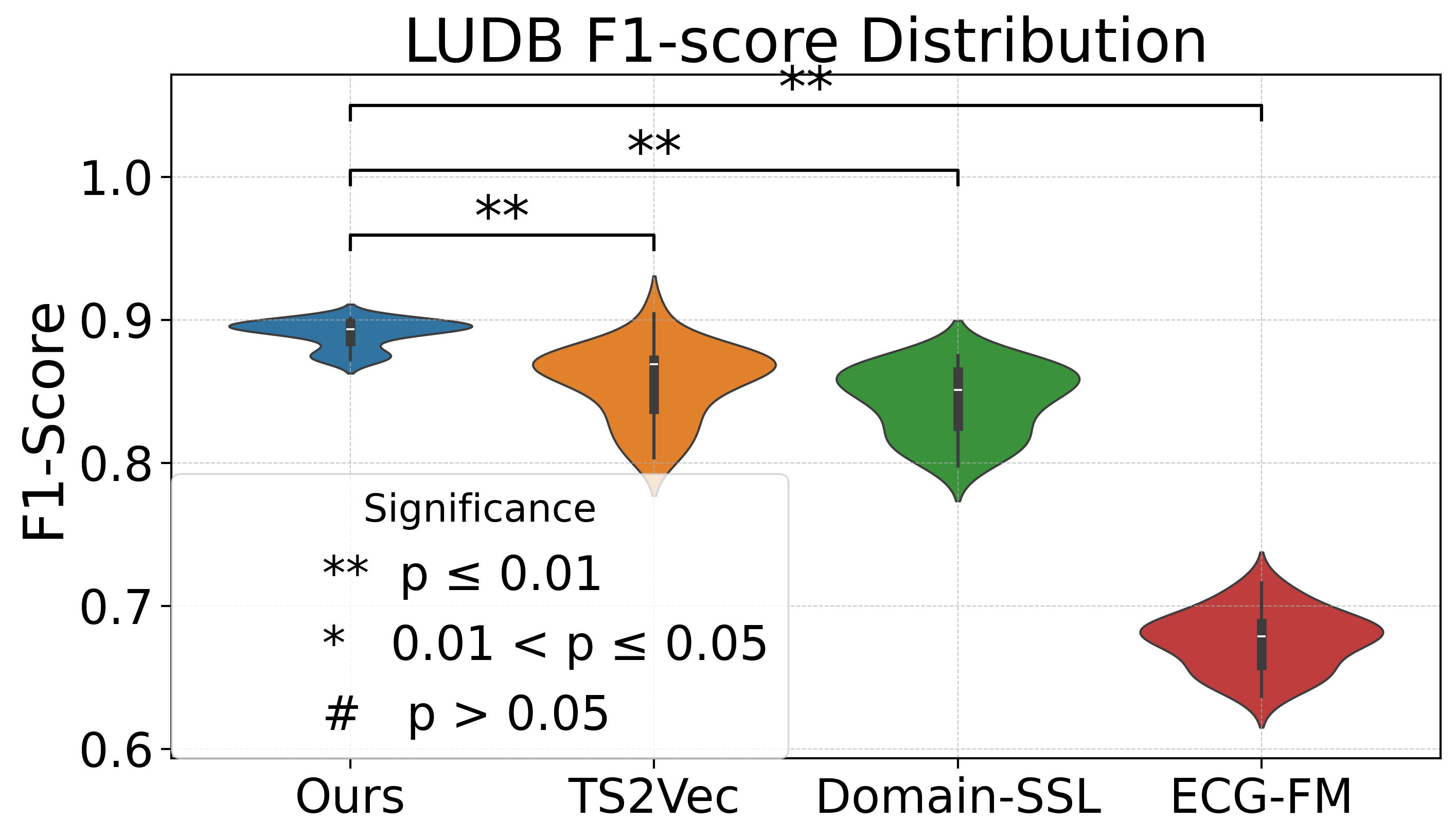}}
  \end{subcaptionbox}
  \hfill
  \begin{subcaptionbox}{LUDB Dice-score\label{fig:signi_lu_dice}}[0.32\textwidth]
    {\includegraphics[width=\linewidth]{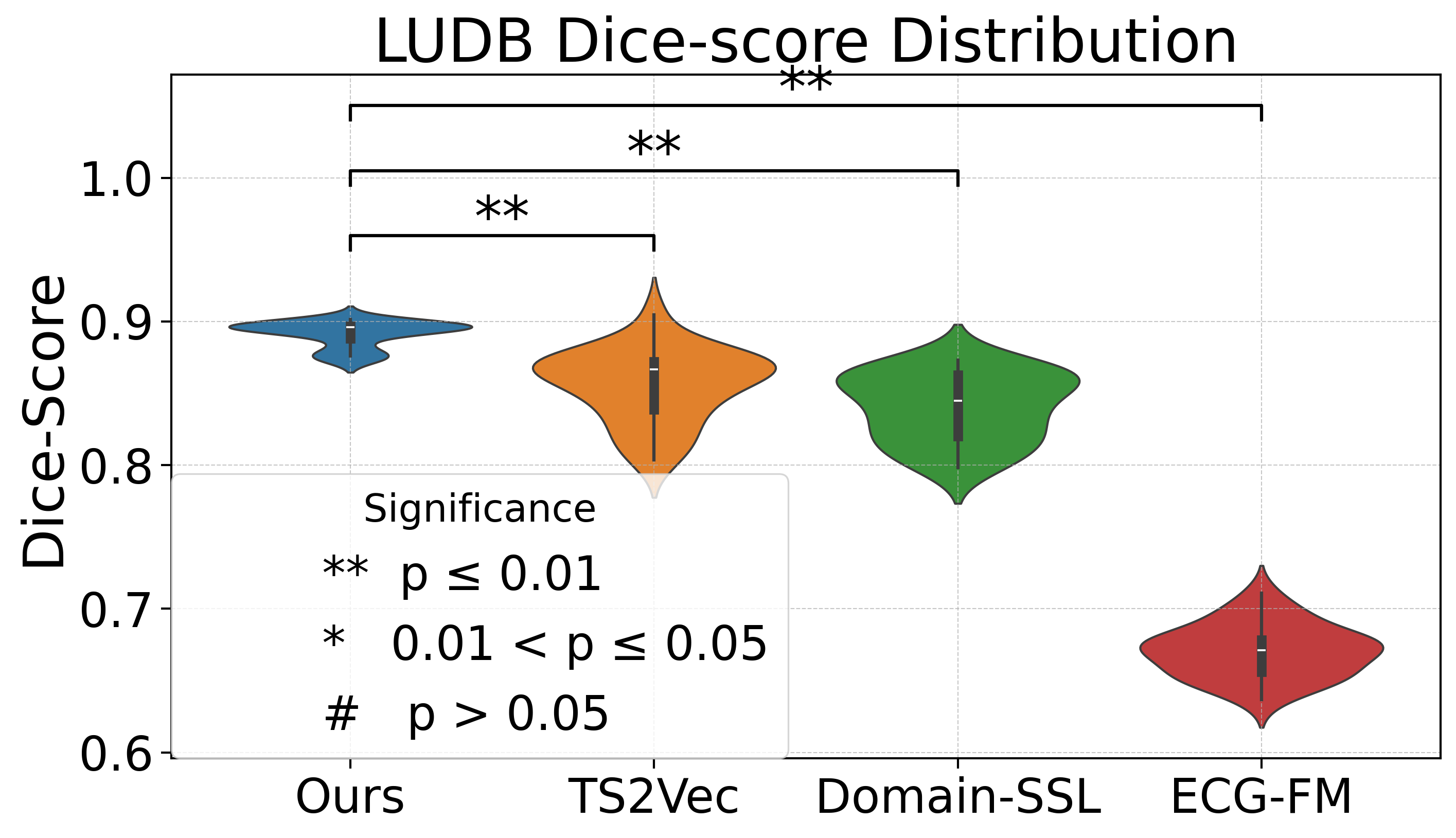}}
  \end{subcaptionbox}
  \caption{Score distributions and Wilcoxon signed-rank test against other methods.}
  \label{fig:significance_test}
\end{figure*}

\section{Results}
\label{sec:results}
To evaluate the effect of each component, an ablation study was conducted with 14 combinations of rhythm level contrasting, beat-level contrasting, and exponentiating on both downstream tasks. The result of the ablation study are presented in Table \ref{table:ablation}.

The combination of the first version of soft contrasting (soft\_1) at the rhythm level, hard contrasting at the beat level, and exponentiating with a power of 50 achieved the best overall performance for both tasks. The second-best combination for task 1 was soft\_1 with exponentiation with a power of 50 applied at both levels, while for task 2, the best alternative was soft\_1 at the rhythm level with no contrasting at the beat level.

Component-wise, analysis reveals that using an exponent of 50 improves performance at all tasks and configurations involving the first contrasting method. At the rhythm level, incorporating soft\_1 enhances performance for both downstream tasks, while soft\_2 improves results for task 2 but reduces performance of task 1. At the beat level, it is challenging to isolate the effect because it heavily depends on other configurations to achieve the best results.

\begin{table}[]
\caption{Results of ablation study on tasks 1 and 2.}
\label{table:ablation}
\centering
\begin{tabular}{ccc||cr|r}
\multicolumn{3}{c||}{\textbf{Params}}                    & \multicolumn{2}{c|}{\textbf{Task 1}}                    & \multicolumn{1}{c}{\textbf{Task 2}} \\
\hline
\textbf{Rhyt C.} & \textbf{Beat C.} & \textbf{Exp.} & \textbf{AUROC}       & \multicolumn{1}{c|}{\textbf{F1}} & \multicolumn{1}{c}{\textbf{F1}}     \\
\hline \hline
hard\textsuperscript{*}               & -                  & 1            & 0.811                & 0.595                           & 0.802                               \\
hard               & hard               & 1            & 0.797                & 0.551                           & 0.827                               \\
hard               & soft\_1            & 1            & 0.789                & 0.552                           & 0.785                               \\
hard               & soft\_1            & 50           & 0.794                & 0.540                           & 0.805                               \\
hard               & soft\_2            & 1            & 0.795                & 0.567                           & 0.784      \\
soft\_1            & -                  & 1            & {\ul 0.834}          & 0.593                           & 0.869                               \\
soft\_1            & -                  & 50           & 0.814                & 0.572                           & {\ul 0.895}                               \\
soft\_1            & hard               & 1            & 0.681                & 0.247                           & 0.784                               \\
soft\_1            & hard               & 50           & {\ul \textbf{0.843}} & {\ul \textbf{0.622}}            & {\ul \textbf{0.911}}                \\
soft\_1            & soft\_1            & 1            & 0.704                & 0.389                           & 0.807                               \\
soft\_1            & soft\_1            & 50           & 0.833                & {\ul 0.598}                     &  0.881                         \\
soft\_2            & -                  & 1            & 0.797                & 0.534                           & 0.888                               \\
soft\_2            & hard               & 1            & 0.801                & 0.525                           & 0.872                               \\
soft\_2            & soft\_2            & 1            & 0.805                & 0.537                           & 0.878                              \\
\hline
*baseline
\end{tabular}
\end{table}

We compared our results with other frameworks, including TS2Vec, Domain-SSL, and ECG-FM. The original ECG-FM was trained on MIMIC-IV\cite{johnsonMIMICIVFreelyAccessible2023}, PhysioNet2021\cite{reynaWillTwoVarying2021}, and UHN datasets. However, due to privacy concerns, the authors released only the weights pretrained on publicly available datasets. For our model, we used the best configuration described in the previous paragraph. To ensure a fair comparison, all methods except for ECG-FM were pretrained on the same folds of the PTB-XL dataset. Identical downstream training configurations were applied across all methods, including the same linear probing layer and decoder architecture for task 1 and task 2, respectively. This approach allows us to evaluate performance by leveraging the strengths of the contrastive learning framework while minimizing external influences. We performed 5 runs of k‑fold cross‑validation for each model, using $k=10$ for task 1 and $k=5$ for task 2. For task 1, only folds 9 and 10 were used for testing to avoid any data leakage from pretraining. Finally, we performed a Wilcoxon signed‑rank test with Bonferroni correction to assess statistical significance, following the recommendation in \cite{rainioEvaluationMetricsStatistical2024}, as shown in Fig.-\ref{fig:significance_test}

In task 1, ECG-FM achieved the best performance, showing statistically significant improvement over other methods. We expected this results, as ECG-FM was pretrained on \textpm 700,000 data points, which is $31.8\times$ more than the other methods. Our method achieves the second-best result, with strong evidence ($p < 0.01$) against TS2Vec and Domain-SSL. Notably, we reached 93\% of the best F1-score (Fig.-\ref{fig:significance_test}.A) using $31.8\times$ less data during pretraining. In task 2, our method achieves the best results in both F1-score and Dice-score, with strong evidence against all other models ($p < 0.01$).

It achieved 4\% higher score (Fig.-\ref{fig:significance_test}.B) than the second-best, TS2Vec leverages temporal contrasting at multiple levels to capture local context. These results suggest that by utilising soft contrast at the rhythm level and hard contrasting at the beat level, our method effectively captures local context information. This competitive performance at both the global and local levels may be beneficial in classifying conditions that requires analysis of both rhythm and morphological features. Providing a model with a strong local representation can significantly improve the classification in such cases.

\section{Conclusion}
\label{sec:conclusion}
We demonstrated that contrastive learning can effctively improve performance on two critical 12-lead ECG analysis tasks: multilabel classification and ECG wave segmentation. Our novel framework, which leverages exponentiated soft targets derived from feature similarity for rhythm contrasting and heartbeat contrasting, achieves particularly strong results in the segmentation task. This superior performance suggests that our approach successfully captures fine-grained local contextual information within the ECG signal, which is essential for precise wave boundary delineation.
%This result suggests that this framework is more effective at capturing the local contextual information of the ECG signal.

\vspace{-1mm}
\section{Compliance with ethical standards}
\label{sec:ethics}
This research study was conducted retrospectively using human subject data made available in open access by PTB-XL and LUDB datasets. Ethical approval was not required as confirmed by the license attached with the open access data.

\vspace{-1mm}
\section{Acknowledgments}
\label{sec:acknowledgments}
M.I.R. is funded by Endowment Fund for Education (LPDP) scholarship, Ministry of Finance, Republic of Indonesia. F.D acknowledges funding from EPSRC (Grant No.~EP/W01212X/1) and the Academy of Medical Sciences (Grant No.~NGR1/1678).

% References should be produced using the bibtex program from suitable
% BiBTeX files (here: strings, refs, manuals). The IEEEbib.bst bibliography
% style file from IEEE produces unsorted bibliography list.
% ------------------------------------------------------------------------- 
\vspace{-1mm}
% \bibliography{references}

\bibliographystyle{ieeetr}
\bibliography{references}  %%% Uncomment this line and comment out the ``thebibliography'' section below to use the external .bib file (using bibtex) .

\end{document}